\documentclass[letterpaper, 10pt, conference]{ieeeconf}
\IEEEoverridecommandlockouts  

\usepackage{amsmath,amssymb,amsfonts}
\usepackage{algorithmic}
\usepackage{graphicx}
\usepackage{textcomp}
\usepackage{xcolor}

\usepackage{multicol}
\usepackage[bookmarks=true]{hyperref}
\usepackage{bm}
\usepackage{amsmath}
\usepackage{amssymb}
\usepackage{xcolor}

\usepackage{graphicx}
\usepackage{booktabs}
\usepackage{multirow}
\usepackage{placeins}
\usepackage{soul}
\usepackage{subfig}
\usepackage{mathtools}
\usepackage{makecell}

\begin{document}

\title{\textbf{Predictive Planner for Autonomous Driving with Consistency Models}}

\author{
  Anjian Li$^{*1,2}$,
  Sangjae Bae$^{1}$,
  David Isele$^{1}$,
  Ryne Beeson$^{2}$,
  Faizan M. Tariq$^{*1}$
  \thanks{$^{1}$Honda Research Institute, San Jose, CA, 95134, USA}
  \thanks{$^{2}$Princeton University, Princeton, NJ, 08544, USA}
  \thanks{$^{*}$Corresponding authors: {\tt anjianl@princeton.edu} \& \newline {\tt faizan\_tariq@honda-ri.com}}
}

\maketitle

\begin{abstract}

Trajectory prediction and planning are essential for autonomous vehicles to navigate safely and efficiently in dynamic environments.
Traditional approaches often treat them separately, limiting the ability for interactive planning.
While recent diffusion-based generative models have shown promise in multi-agent trajectory generation, their slow sampling is less suitable for high-frequency planning tasks.
In this paper, we leverage the consistency model to build a predictive planner that samples from a joint distribution of ego and surrounding agents, conditioned on the ego vehicle's navigational goal.
Trained on real-world human driving datasets, our consistency model generates higher-quality trajectories with fewer sampling steps than standard diffusion models, making it more suitable for real-time deployment.
To enforce multiple planning constraints simultaneously on the ego trajectory, a novel online guided sampling approach inspired by the Alternating Direction Method of Multipliers (ADMM) is introduced.
Evaluated on the Waymo Open Motion Dataset (WOMD), our method enables proactive behavior such as nudging and yielding, and also demonstrates smoother, safer, and more efficient trajectories and satisfaction of multiple constraints under a limited computational budget.
The project website is at \href{https://anjianli21.github.io/projects/predictive_planner/}{https://anjianli21.github.io/projects/predictive\_planner/}.

\end{abstract}


\section{Introduction}


\begin{figure*}[t]
    \centering
     \vspace{0.5em}
\includegraphics[width=0.7\textwidth]{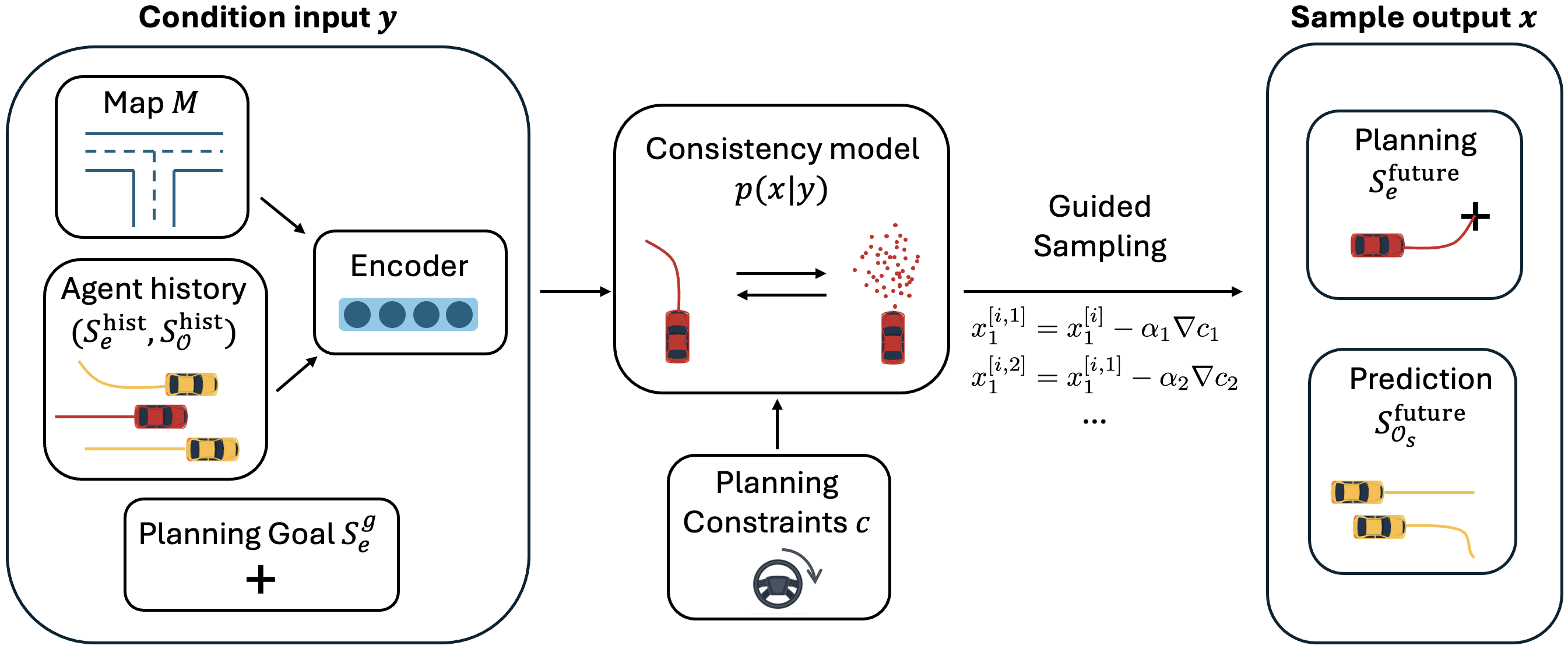}
    \caption{Our consistency model takes the map and agents history encoded with the planning goal, and generates a trajectory plan for the ego vehicle and predictions for surroundings. Multiple constraints are enforced through online guided sampling.}
    \label{fig:method workflow}
\end{figure*}

To navigate safely and efficiently in a dynamic environment, autonomous vehicles must effectively predict and interact with diverse road participants, including other vehicles, pedestrians, and cyclists.
This typically requires a prediction module to anticipate other agents' future trajectories and a planning module to generate trajectories for the ego vehicle.

Traditionally, these two modules usually operate in a decoupled, alternating manner~\cite{chen2023interactive,li2021prediction,tariq2022slas,bidirectional_overtaking}, which introduces fundamental limitations.
Some approaches integrate data-driven predictions into an optimization-based planner that is inherently reactive rather than proactive~\cite{gupta2023interaction}.
In highly interactive scenarios, such as lane merging where proactive nudging is essential, reactive planning may cause the ego vehicle to be stuck \cite{bae2020cooperation, unstuck}.
Also, existing work attempts to incorporate the ego vehicle's plan into the prediction module~\cite{huang2023conditional}, but the initial ego plans may already be suboptimal if ignoring future interaction with others.
In addition, the alternation between prediction and planning introduces computational inefficiency, making it challenging to meet the demands of real-time operation.

Recently, there has been growing interest in multi-agent trajectory generation using purely data-driven approaches.
Transformer-based methods have been widely adopted for multi-agent prediction tasks~\cite{shi2024mtr++,ngiam2021scene}.
Diffusion models, with their strong ability to sample from high-dimensional multimodal distributions, have been adopted that improve the prediction performance~\cite{jiang2023motiondiffuser}.
Compared to transformer-based models, diffusion models also support controllable generation through guidance methods, making them particularly suitable for scenario generation \cite{huang2024versatile,zhong2023guided,chang2023controllable}.

However, diffusion models typically require many sampling steps to obtain high-quality samples, which becomes challenging to meet the demands of real-time operations on autonomous vehicles.
To address this limitation, consistency models~\cite{song2023consistency} are designed to generate data directly from noise in a single step or a small number of steps.
This significantly improves sample efficiency and maintains high quality under limited computational budgets\cite{ding2023consistency, prasad2024consistency}.

In this paper, we present a predictive planner for autonomous driving with a unified data-driven framework.
Our key idea is to leverage a consistency model to sample the joint distribution of the ego and surrounding agents' future trajectories, given the map, trajectory history, and the ego vehicle's goal. 
Consistency models enable high-quality trajectory generation with only a few sampling steps, making them more suitable for online planning than standard diffusion models.
Trained on WOMD \cite{ettinger2021large}, our method captures complex agent interactions and produces proactive trajectories observed in real-world driving.
To further enforce multiple planning constraints, we introduce a novel online guided sampling approach inspired by ADMM~\cite{boyd2011distributed}.
Our planner is also encoder-agnostic and can be integrated with any suitable scene encoder.

Experiments show that our consistency model-based planner generates proactive behaviors, e.g., nudging and yielding, without alternation between prediction and planning.
It outperforms both diffusion models and an accelerated variant DDIM~\cite{song2020denoising} with smoother, safer, and more efficient trajectories under limited computational budgets.
Compared to transformer-based models, we enable controllable generation that significantly improves satisfaction of multiple trajectory constraints without affecting the original distribution.

\section{Related Work}

Traditionally, trajectory planning is often formulated as a numerical optimization problem, typically decomposed into path planning and speed planning to facilitate real-time operation in dynamic environments~\cite{anon2024multi, fcp}.
Beyond standalone planning, preliminary effort have integrated trajectory predictions into planner design for collision avoidance~\cite{li2021prediction, rcms, multifuture, nawaz2025graph} and proactive motion planning \cite{bae2022lane, activeProbing, iann-mppi, saxena2020driving}, but either struggle to scale with the number of agents \cite{isele2019interactive} or compromise optimality due to a restrictive exploration of the solution search space \cite{song2024efficient}.
Game-theoretic approaches have been explored to model agent interactions~\cite{sadigh2016planning}, with efficient computation~\cite{fisac2019hierarchical} that considers imperfect traffic agents~\cite{tian2021anytime}, safe actions~\cite{tian2022safety}, and uncertainty reduction in the estimates of agent behaviors~\cite{hu2024active}.

Recently, data-driven methods have shown promising results in trajectory predictions when trained on large-scale real-world human driving datasets such as Waymo Open Motion Dataset (WOMD)~\cite{ettinger2021large}, nuScenes~\cite{caesar2020nuscenes}, etc.
Transformer~\cite{vaswani2017attention} based methods, such as Scene Transformer~\cite{ngiam2021scene} and Motion Transformers (MTR)~\cite{shi2022motion}, use an encoder-decoder architecture and achieved great performances on Waymo Motion Prediction Challenge~\cite{waymo2024motion}.
They allow for multi-agent trajectory generation~\cite{ngiam2021scene,shi2024mtr++} with potential for goal-conditioned trajectory planning.
However, trajectory planning usually requires more controllable generation to satisfy constraints provided online, which remains challenging for transformer-based approaches.

Diffusion models are powerful generative models capable of producing high-quality samples from complex distributions, initially gaining success in computer vision~\cite{sohl2015deep,song2020score,ho2020denoising}.
The Denoising Diffusion Probabilistic Model (DDPM)~\cite{ho2020denoising} corrupts training data with Gaussian noise and learns to reverse this process through iterative denoising.
Controllable generation in diffusion models can be achieved through classifier guidance~\cite{dhariwal2021diffusion} during sampling.
As a result, Diffusion models have been applied to robot policy learning~\cite{janner2022planning,ajay2022conditional,chi2023diffusion} and trajectory generation~\cite{li2024diffusolve,li2025aligning,zhang2024predicting}.
In autonomous driving, diffusion model facilitates multimodal trajectory predictions~\cite{jiang2023motiondiffuser} and controllable traffic scenario generation~\cite{zhong2023guided,chang2023controllable,huang2024versatile}, capturing complex interactive behaviors.

However, diffusion models typically require a large number of sampling steps to maintain sample quality, resulting in substantial computational overhead.
To address this inefficiency, Denoising Diffusion Implicit Models (DDIM)~\cite{song2020denoising} introduces a deterministic sampling process that balances speed and quality, while progressive distillation~\cite{salimans2022progressive} reduces the number of sampling steps by distilling a pre-trained diffusion model into a more efficient one.
Consistency model~\cite{song2023consistency,song2023improved} takes a different approach by directly generating
data from noise in a single or a few steps.
They outperform distillation methods~\cite{song2023consistency} in image generation and have shown early success in robot policy learning~\cite{ding2023consistency,prasad2024consistency}.



In this work, we focus on building a goal-conditioned predictive planner using a consistency model.
Unlike offline scenario generation, where sampling efficiency is less critical, our method produces high-quality interactive behaviors with significantly fewer sampling steps than diffusion models, making it well-suited for high-frequency online planning.
Compared to diffusion-based predictors, which model the full multimodal future distribution, our approach focuses on goal-conditioned generation tailored for planning.
In contrast to transformer-based models, which lack controllable generation at inference time, our approach enables multiple planning constraints, e.g., goal-reaching and control limits, to be satisfied simultaneously through an ADMM-inspired online guided sampling.
\section{Problem Formulation} \label{sec: problem formulation}

Assuming an ego vehicle's goal is given from a high-level planner and the trajectory histories of all traffic agents and map information are known, we aim to plan the trajectory of the ego vehicle while simultaneously predicting the interactive behaviors of others.

At the current timestep $t_0 \in \mathbb{R}$, the ego agent's state history over the past $H_1 \in \mathbb{N}$ timesteps including the current state 
is given as $S_e^{\text{hist}} = (s_e^{t_0}, s_e^{t_0-1}, ..., s_e^{t_0-H_1}) \in \mathbb{R}^{(H_1 + 1) \times d_s}$, where $d_s$ is the dimension of the states, e.g. position, velocity, etc.
Let $\mathcal{O}$ denote the set of 
other agents of all types, such as
vehicles, pedestrians, and cyclists.
Their history states are given for the current timestep $t_0$ and the past $H_1$ timesteps, denoted by $S_{\mathcal{O}}^{\text{hist}} = (s_{o}^{t_0}, s_{o}^{t_0-1}, ..., s_{o}^{t_0-H_1})_{o \in \mathcal{O}} \in \mathbb{R}^{N_{\mathcal{O}} \times (H_1 + 1) \times d_s}$, where $N_{\mathcal{O}} = |\mathcal{O}|$.
The map information is denoted by $M \in \mathbb{R}^{N_m \times c_p \times d_m}$, which includes $N_m \in \mathbb{N}$ polylines in the scenario with $c_p$ points on each polyline and $d_m$ attributes per point.
We pad and mask the invalid entries. 
We assume the ego agent's goal state is given as $s_e^g \in \mathbb{R}^{d_s}$.

Let $\mathcal{O}_s \subset \mathcal{O}$ denote the subset of surrounding agents that may interact with the ego agent.
Given the information $(S_e^{\text{hist}}, S_{\mathcal{O}}^{\text{hist}}, M, s_e^g)$, for the future $H_2 \in \mathbb{N}$ timesteps, we aim to plan the ego agent's trajectory $S_e^{\text{future}} = (s_e^{t_0+1}, s_e^{t_0+2}, ..., s_e^{t_0+H_2}) \in \mathbb{R}^{H_2 \times d_s}$, and predict the trajectories of surrounding agents $S_{\mathcal{O}_s}^{\text{future}} = (s_{o}^{t_0+1}, s_{o}^{t_0+2}, ..., s_{o}^{t_0+H_2})_{o \in \mathcal{O}_s} \in \mathbb{R}^{N_{\mathcal{O}_s} \times H_2 \times d_s}$, where $N_{\mathcal{O}_s} = |\mathcal{O}_s|$.
This joint prediction and planning naturally captures interactive behaviors between agents.
Additionally, the planned trajectory for the ego vehicle should satisfy $N_c \in \mathbb{N}$ constraints $c_i( S_e^{\text{future}}) \leq 0, i=1,2,...,N_c$, which may include goal reaching, control limits, or other requirements.

\section{Preliminaries}

\subsection{Motion Transformer Encoder} \label{sec: mtr encoder}

To encode the trajectory history $(S_e^{\text{hist}}, S_\mathcal{O}^{\text{hist}})$ and map information $M$ as the conditional input for the consistency model, we adopt the encoder architecture from MTR~\cite{shi2022motion,shi2024mtr++}.
This transformer-based encoder effectively models scene context and agents' interactions, and introduces a dense prediction head with a loss function $\mathcal{L}_{\text{encoder}}$~\cite{shi2022motion} to train this encoder individually.
The MTR++ encoder \cite{shi2024mtr++} would be suitable for multi-agent settings, but its implementation is not yet available.
More importantly, our method is encoder-agnostic, allowing for flexible choice of any suitable scene encoder to be integrated.
We represent this encoded feature of the trajectory history and map as $\text{MTR}(S_e^{\text{hist}}, S_\mathcal{O}^{\text{hist}}) \in \mathbb{R}^{(N_{\mathcal{O}} + 1) \times D}$ and $\text{MTR}(M) \in \mathbb{R}^{N_m \times D}$, respectively, where $D$ is the embedded feature dimension.

\subsection{Consistency Model}\label{sec: consistency model intro}

The consistency model can generate high-quality samples from complex distributions with only one or a few sampling steps~\cite{song2023consistency}.
It consists of a forward diffusion process and a reverse process.
Let $p_{\text{data}}$ be the data distribution with zero mean and unit variance over space $\mathcal{X}$.
In the forward process, we first draw an initial sample $x_1$ from $p_{\text{data}}$, then gradually corrupt it with an increasing noise schedule ${\sigma_1, \sigma_2, ..., \sigma_T}$ through $T$ steps,  with $\sigma_1 \approx 0$.
At each step $i$, the corrupted data is $x_i = x_1 + \sigma_i \cdot \epsilon$  with $\epsilon$ sampled from $\mathcal{N}(0, I)$.
$\sigma_T$ is chosen large enough such that the empirical distribution of the resulting $x_T$ approximates $\mathcal{N}(0, \sigma_T^2 I)$.

Let $\mathcal{Y}$ be the space of conditioning information.
In the reverse process, we aim to learn a consistency function $f_{\theta}$ with the parameter $\theta$ that maps a noisy trajectory sample $x_i$, condition $y \in \mathcal{Y}$, and the noise level $\sigma_i$ directly to the corresponding clean sample $x_1$ for each step $i=1,..,T$.
This is achieved by choosing a specific function form of $f_\theta \coloneqq c_{\text{skip}}(\sigma) x + c_{\text{out}}(\sigma) F_\theta(x, y, \sigma)$,
where $c_{\text{skip}}$ and $c_{\text{out}}$ are differentiable function with $c_{\text{skip}}(\sigma_1)=1, c_{\text{out}}(\sigma_1)=0$~\cite{karras2022elucidating}, such that the boundary condition is met: $f_{\theta}(x_1, y, \sigma_1) = x_1$.
$F_{\theta}$ is usually approximated by neural networks.

For the consistency model training, we aim to enforce the consistency of the output from $f_\theta$ for adjacent sampling steps $i$ and $i+1$ to be both $x_1$,
through the following loss $\mathcal{L}_{\text{consistency}}$:
\begin{align}\label{eq: consistency training loss}
    &\mathcal{L}_{\text{consistency}} =  \nonumber \\ &\mathbb{E}_{i \sim \mathcal{P}[1,T-1], \epsilon \sim \mathcal{N}(0, I)} \bigl[d(f_{\theta}(x_i, y, \sigma_i) - f_{\theta}(x_{i+1}, y, \sigma_{i+1}))\bigr]
\end{align}
where $\mathcal{P}[1,T-1]$ is lognormal distribution over the integer set $\{1, 2, ..., T-1\}$, and $d$ is a pseudo-huber metric function $d(x,y) = \sqrt{||x-y||_2^2 + \delta^2} - \delta$~\cite{song2023improved} where $\delta$ is a small constant.

For data generation, we draw a sample $x_T$ from $\mathcal{N}(0, \sigma_T^2 I)$.
Then with the trained consistency model $f_{\theta}$, we perform iterative sampling by first predicting the approximate clean data $x_1^{[i]}$ and then sample $x_{i-1}$ with $\epsilon$ drawn from $\mathcal{N}(0, I)$:
\begin{align} \label{eq: consistency sampling}
    x_1^{[i]} = f_{\theta}(x_i, y, \sigma_i), \quad x_{i-1} = x_1^{[i]} + \sigma_{i-1} \epsilon
\end{align}
for $i = T, T-1, ..., 1$, until we obtain the clean sample $x_1^{[1]}$.

\section{Methodology}


Our method is summarized in Fig. \ref{fig:method workflow}.
We first use an MTR encoder to encode the agent's trajectory history and map information.
With the ego agent's planning goal and the MTR-encoded features, a consistency model generates trajectory plans and predictions for the ego and surrounding agents, respectively.
Multiple planning constraints for the ego agent are achieved through ADMM-inspired online guided sampling.


\subsection{Data pre-processing}\label{sec: data preprocessing}

Instead of using an ego-centric coordinate frame, we apply a coordinate transformation $\Gamma$ similar to MTR++ ~\cite{shi2024mtr++} that maps each agent's trajectory into its own local coordinate frame, which is centered at each agent's position at the current timestep $t_0$.
The transformed data $(\Gamma(S_e^{\text{future}}),\Gamma(S_{\mathcal{O}_s}^{\text{future}})) \in \mathbb{R}^{(N_{\mathcal{O}_s} + 1) \times H_2 \times d_s}$ has substantially reduced variance.
We then standardize the transformed trajectories to zero mean and unit variance.
To preserve the relative spatial relationships among agents, the reference states are collected as $\text{Ref}(S_e^{\text{future}}, S_{\mathcal{O}_s}^{\text{future}}) \in \mathbb{R}^{(N_{\mathcal{O}_s} + 1) \times d_s}$ consisting of each agent's own position at timestep $t_0$.

\subsection{Consistency Model Training}

\begin{figure*}[t]
    \centering
    \vspace{0.5em}
    \subfloat[A proactive nudging behavior from the ego car to switch a lane.]{
        \includegraphics[width=0.75\textwidth]{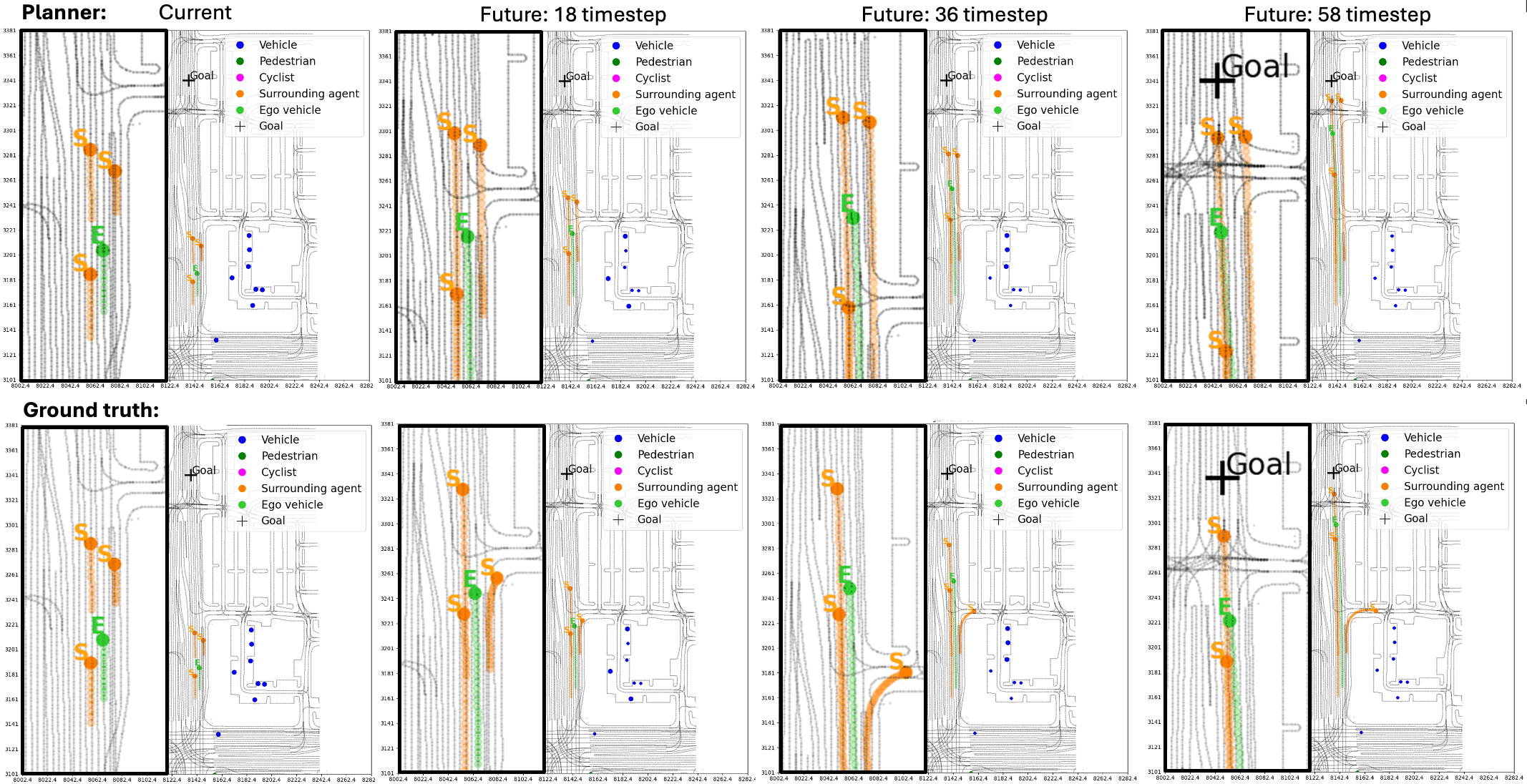}
        \label{fig:nudging}
    }\\[0.5em] 
    \subfloat[A safe yielding behavior for the ego car to make a turn.]{
        \includegraphics[width=0.75\textwidth]{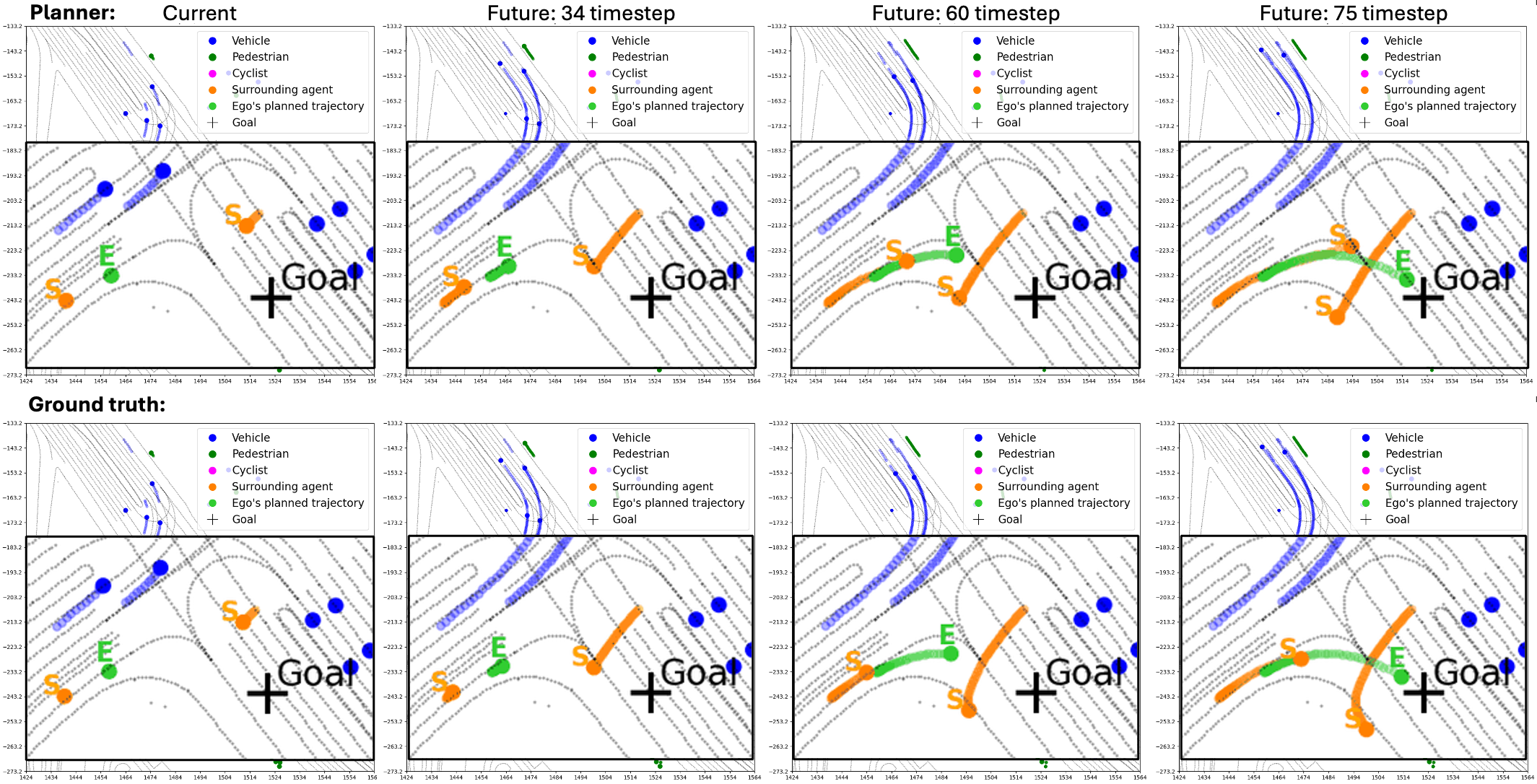}
        \label{fig:yielding}
    }
    \caption{For each scenario, upper row: interactive behavior from our predictive planner; 
    lower row: Ground truth driving behaviors.
    The ego agent's trajectory is in green annotated by ``E" and the surrounding agents' trajectories are in orange annotated by``S".
    The other agents' trajectories (always shown with ground truth) are in blue.}
    \label{fig: interactive behavior}
\end{figure*}

Let $\mathcal{X}$ denote the space of transformed future trajectories $\Gamma(S_e^{\text{future}})$ and $\Gamma(S_{\mathcal{O}_s}^{\text{future}})$ from Sec. \ref{sec: data preprocessing}. 
Let $\mathcal{Y}$ denote the space of conditional inputs, containing encoded historical trajectories $\text{MTR}(S_e^{\text{hist}}, S_{\mathcal{O}}^{\text{hist}})$, map features $\text{MTR}(M)$ from Sec. \ref{sec: mtr encoder}, ego agent's planning goal $s_e^g$, and the reference coordinates $\text{Ref}(S_e^{\text{future}}, S_{\mathcal{O}_s}^{\text{future}})$ from Sec. \ref{sec: data preprocessing}.
Given a sample $y \in \mathcal{Y}$,  we leverage the consistency model \cite{song2023consistency,song2023improved} to sample future trajectories from the conditional probability distribution $p(\cdot|y)$ over $\mathcal{X}$.
Each trajectory sample represents one possible joint future behavior of the ego vehicle and surrounding agents.


For training our predictive planner, we jointly train the MTR encoder and the consistency model together with a hybrid loss function, which is a weighted sum of the consistency training loss $\mathcal{L}_{\text{consistency}}$ in Eq. \eqref{eq: consistency training loss} and the dense prediction loss $\mathcal{L}_{\text{encoder}}$ of the MTR encoder from Sec. \ref{sec: mtr encoder} with details in~\cite{shi2022motion} as follows:
\begin{align}
    \mathcal{L} = \omega_0 \mathcal{L}_{\text{consistency}} + \omega_1 \mathcal{L}_{\text{encoder}}
\end{align}
where $\omega_0$ and $\omega_1$ control
the weight of these two losses.

\subsection{Planning Constraints Construction} \label{sec: planning constraints}
To make the sampled trajectories suitable and smooth for planning use, we apply several constraints on the ego trajectories.
Assuming the vehicle has the following dynamics
\begin{align} \label{eq: dynamics equation}
    \dot p_x = v \cos \theta, \quad \dot p_y = v \sin \theta, \quad
    \dot v = a, \quad \dot \theta = \omega
\end{align}
where $p_x$ and $p_y$ are positions, $v$ is the linear speed and $\theta$ is the orientation.
The control input is the acceleration $a$ and angular speed $\omega$.
We only use $p_x$ and $p_y$ from the future states $S_e^{\text{future}}$ to construct our planning constraint function $c_i$, since other states like $v$ or $\theta$ in the dataset may not satisfy the dynamics equations in Eq. \eqref{eq: dynamics equation} and are also quite noisy.
Using the differential flatness for dynamics in Eq. \eqref{eq: dynamics equation}, we infer 
the control input $a$ and $\omega$ as $a = \frac{\dot p_x \ddot p_x + \dot p_y \ddot p_y}{\sqrt{\dot p_x^2 + \dot p_y^2}}, \omega = \frac{\dot p_x \ddot p_y - \dot p_y \ddot p_x}{\dot p_x^2 + \dot p_y^2}$
through finite differencing with inevitable noises.

With $p_x, p_y$ and the derived $a$ and $\omega$, we consider three types of planning constraints $c$ to minimize as follows.
\begin{align} \label{eq: 3 planning constraints}
    &c_{\text{goal}} = \sqrt{(p_x^{t_0 + H_2} - p_{x, goal})^2 + (p_y^{t_0 + H_2} - p_{y, goal})^2} \nonumber \\
    &c_{\text{acc}} = \frac{1}{H_2} \sum_{i=t_0+1}^{t_0 + H_2} max(|a^i| - a_{\text{limit}}, 0) \nonumber \\
    &c_{\omega} = \frac{1}{H_2} \sum_{i=t_0+1}^{t_0 + H_2} max(|\omega^i| - \omega_{\text{limit}}, 0)
\end{align}
$c_{\text{goal}}$ enforces goal-reaching for the ego vehicle 
at the final timestep $t_0 + H_2$.
$c_{\text{acc}}$ and $c_{\omega}$ are control limit constraints over the ego agent's trajectory for a given $a_{\text{limit}}$ and $\omega_{\text{limit}}$.
These constraints are inherently in conflict. 
For example, only moving the trajectory’s final state closer to the goal can increase acceleration violations of the car.

\subsection{Guided Sampling} \label{sec: guided sampling}

To simultaneously enforce multiple planning constraints, like in Eq.~\eqref{eq: 3 planning constraints}, on the trajectory samples without affecting the original distribution, we present a novel ADMM-inspired guided sampling approach, which is applied solely during sampling at test time, without requiring any modification to the training procedure.


Assuming there are $N_c \in \mathbb{N}$ planning constraint $c_j$ to minimize, $j = 1,2, ..., N_c$, standard classifier-guidance~\cite{dhariwal2021diffusion} performs a vanilla gradient descent step on the predicted $x_1^{[i]}$ during each sampling step $i$ in Eq. \eqref{eq: consistency sampling} as follows:
\begin{align} \label{eq: original gradient descent}
     &x_1^{[i, 1]} = x_1^{[i]} - \sum^{N_c}_{j=1}\alpha_{j} \nabla c_j  
\end{align}
where $\nabla c_j$ represents the first-order gradient of the planning constraint $c_j$ and can be computed using auto-differentiation.
$\alpha_j$ is the corresponding step size for each constraint $c_j$.
Then the tuned $x_1^{[i, 1]}$ is used to sample $x_{i-1}$ in Eq. \eqref{eq: consistency sampling}.


However, optimizing multiple constraints simultaneously with conflicts presents a challenge in finding a suitable stepsize $\alpha$.
To address this, we propose a novel alternating direction method inspired by ADMM~\cite{boyd2011distributed} that sequentially applies gradient step for each constraint during the sampling step $i$:
\begin{align} \label{eq: admm gradient step}
    &x_1^{[i, j+1]} = x_1^{[i,j]} - \alpha_{j} \nabla c_{j}, \quad \text{for} \quad j = 1, ..., N_c 
\end{align}
We find that this alternating update strategy demonstrates effective convergence and empirically leads to lower violations of multiple conflicting constraints, compared to standard guided sampling with vanilla gradient descent in Eq.~\eqref{eq: original gradient descent}. 
It also simplifies step size tuning.
Note that this differs from ADMM since it doesn't have a dual update step.

\section{Experiments}



\subsection{Experiment setup}

\begin{figure}[t]
    \centering
    \vspace{0.5em}
    \includegraphics[width=1.0\columnwidth]{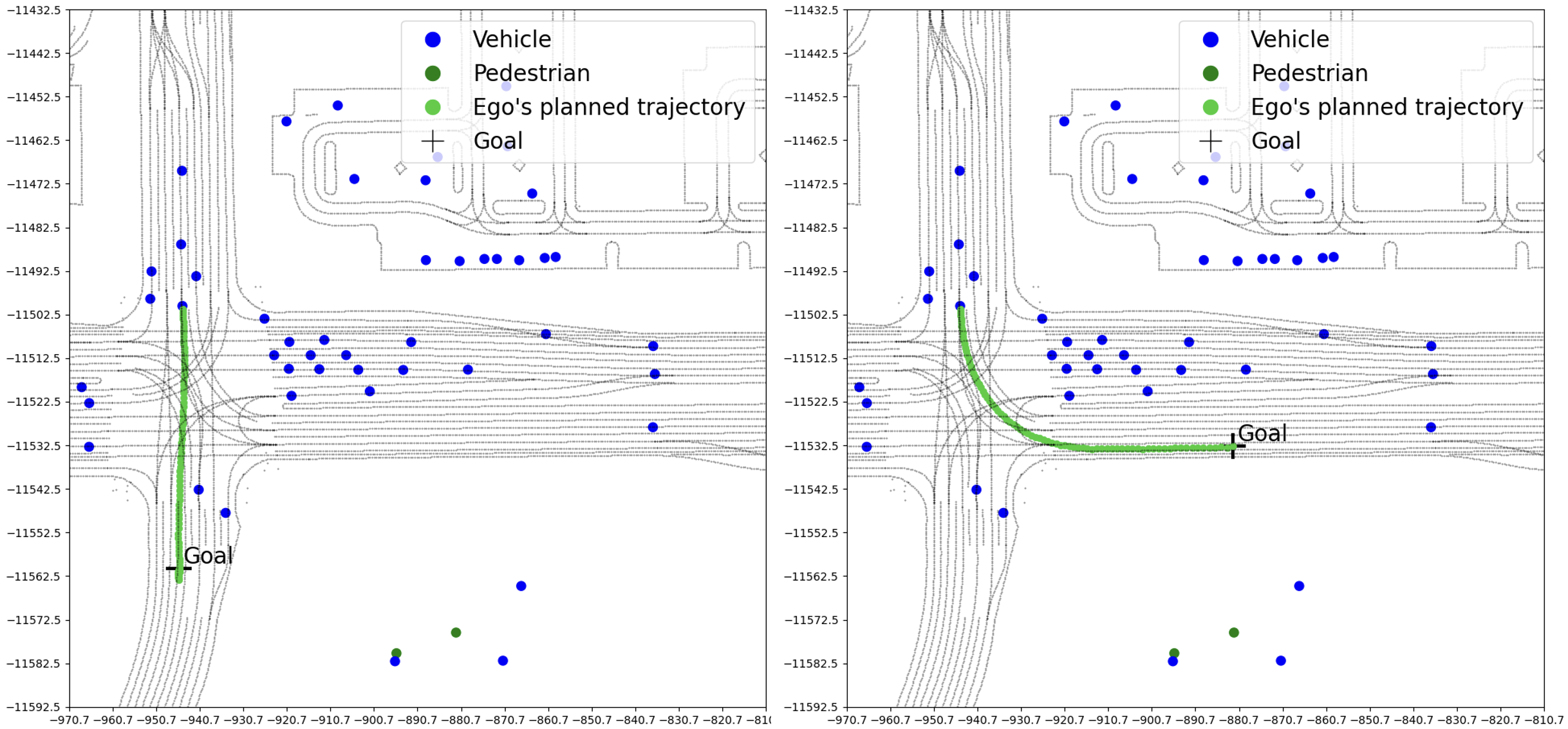}
    \caption{
    Planning with different goal locations using consistency models and guided sampling.}
    \label{fig: different goal}
\end{figure}

\begin{table}[t]
    \centering
    \vspace{0.5em}
    \caption{Planning Trajectory Realism}
    \label{tab:prediction metrics}
    \begin{tabular}{lcccc}
    \toprule
    Method & \makecell{minADE\\(m)} & \makecell{minFDE\\(m)} & \makecell{Sampling\\Step} \\
    \midrule
    MotionDiffuser*~\cite{waymo2024motion} & 0.533 & \textbf{0.007} & 32 \\
    \midrule
    Transformer & 1.237 & 2.796 & 1 \\
    \midrule
    Diffusion-10 & \textbf{0.488} & 0.532 & 10 \\
    Diffusion-4 & 0.545 & 0.711 & 4  \\
    DDIM & 2.749 &  2.304 & 4 \\
    Ours & 0.497 & 0.173 & 4 \\
    Ours (guided) & \textbf{0.490} & \textbf{0.008} & 4 \\
    \bottomrule
    \end{tabular}
\end{table}

\begin{table*}[t]
    \centering
    \vspace{0.5em}
    \caption{Planning Trajectory Quality and Constraint Violations for Different Methods}
    \label{tab:trajectory quality and constraint violation}
    \small
    \setlength{\tabcolsep}{5pt}
    \renewcommand{\arraystretch}{1.2}
    \begin{tabular}{l|cccc|ccc}
    \toprule
    \multirow{3}{*}{\textbf{Method}} 
    & \multicolumn{4}{c|}{\textbf{Planning Trajectory Quality}} 
    & \multicolumn{3}{c}{\textbf{Planning Constraint Violations}} \\
    \cmidrule{2-8}
    & Angle changes & Path length & Curvature & Collision rate & Goal reaching & Acceleration limit & $\omega$ limit \\
    & ($\text{rad/s}$) & ($m$) & ($m^{-1}$) & ($\%$) & ($m$) & ($m/s^{2}$) & ($\text{rad/s}$) \\
    \midrule
    \textbf{Transformer}      & \textbf{0.465} & 55.885 & 0.599 & 15.0 & 8.111  & 0.438   & 1.119 \\
    \midrule
    \textbf{Diffusion-10}     & 0.639 & 67.899 & 1.392 & 1.7 & 2.410  & 35.609  & 2.248 \\
    \textbf{Diffusion-4}      & 0.716 & 61.407 & 1.379 & 2.0 & 3.171  & 19.812  & 2.566 \\
    \textbf{DDIM}             & 0.673 & 135.55 & 1.755 & 18.6 & 14.440 & 175.602 & 4.545 \\
    \textbf{Ours}             & 0.670 & 55.373 & 1.107 & \textbf{0.8} & 0.361  & 4.806   & 2.064 \\
    \textbf{Ours (guided)}    & \textbf{0.487} & \textbf{55.137} & \textbf{0.556} & \textbf{0.8} & \textbf{0.026} & \textbf{0.281} & \textbf{0.519} \\
    \bottomrule
    \end{tabular}
\end{table*}

\begin{table}[t]
\centering
\caption{Increased constraint violation (\%) for vanilla gradient descent compared to the ADMM-inspired method.}
\label{tab:admm_advantage}
\small
\setlength{\tabcolsep}{4pt}
\renewcommand{\arraystretch}{1.0}
\begin{tabular}{@{}lccc@{}}
\toprule
Gradient Step & Goal reaching & Acceleration limit & $\omega$ limit \\
\midrule
50  & 10\% & 26\% & 8\% \\
100 & 9\%  & 35\% & 8\% \\
150 & 12\% & 41\% & 9\% \\
\bottomrule
\end{tabular}
\vspace{-1em}
\end{table}

\subsubsection{\textbf{Dataset}}

We evaluate our method on the Waymo Open Motion Dataset (WOMD)~\cite{ettinger2021large}, which contains 486k traffic scenarios in the training set and 42k scenarios in the interactive validation
set to test on.
For each agent, including vehicles, pedestrians, and cyclists, the dataset includes 1 second of history, current state, and 8 seconds of future trajectory, sampled at 0.1-second intervals.
This sets our history horizon to be $H_1=10$ and future horizons to be $H_2=80$ timesteps.
We identify the surrounding agents with trajectories $S_{\mathcal{O}_s}$ by selecting up to 4 agents closest to the ego vehicle based on their trajectory distances, excluding those beyond a 10-meter threshold.
This choice focuses the predictive planner on the most relevant and interactive neighboring agents, and our model architecture is fully compatible with more agents without additional modifications.
The goal $s_e^g$ of the ego agent is chosen as its state at the end of the 8-second future horizon.

\subsubsection{\textbf{Model Details}}

The MTR encoder consists of 6 transformer layers with a dimension of 256 to encode agent trajectory histories and map information~\cite{shi2022motion}.
The consistency model's conditional embedding uses 3-layer MLPs with 256 neurons per layer.
The consistency function $F_\theta$ is modeled with a U-Net~\cite{ronneberger2015u} with a dimension of 128.

We choose $T=5$, i.e., 4 sampling steps in the forward and backward process for the consistency model.
We set the noise schedule with $\rho = 6$, $\sigma_1 = 0.002, \sigma_T=80$, and $\sigma_i = \left(\sigma_{0}^{1/\rho} + \frac{i-1}{T-1}(\sigma_{T}^{1/\rho} - \sigma_{0}^{1/\rho})\right)^\rho, i = 1, ..., T$ as ~\cite{song2023improved}.
During guided sampling, we perform 100 gradient steps of Eq. \eqref{eq: admm gradient step} at each sampling step, and set the coefficient for $c_{\text{goal}}, c_{\text{acc}}$ and $c_{\omega}$ to be $2\times10^{-5}, 3\times10^{-6}, 5\times10^{-7}$, respectively.

\subsubsection{\textbf{Methods and Baselines}}

We compare our consistency model-based approach against several baselines.
Currently, we provide the evaluation of the open-loop trajectory generation performance of those methods, focusing on interactive behavior through a single inference.
\paragraph{\textbf{Transformer}} This transformer baseline uses MTR~\cite{shi2022motion} to generate trajectory only for ego vehicle.
It has a larger model size for the decoder (512 vs. our 128), but doesn't incorporate the ego vehicle's goal $s_e^g$ as input.
\paragraph{\textbf{Diffusion-4} and \textbf{Diffusion-10}} These diffusion baselines use the same MTR encoder and goal embedding as our approach but adopt DDPM~\cite{ho2020denoising} for trajectory sampling. 
We evaluate two variants: one trained and sampled with 4 steps (Diffusion-4), and another with 10 steps (Diffusion-10).
\paragraph{\textbf{DDIM}}  This baseline aims to accelerate the pretrained Diffusion-10 model with only 4 sampling steps in trajectory generation, with an efficient DDIM sampling strategy~\cite{song2020denoising}.
\paragraph{\textbf{MotionDiffuser}*~\cite{waymo2024motion}} This is state-of-the-art diffusion-based approach for multi-agent trajectory prediction. As the official code has not been released, we report their results of trajectory realism directly from the original paper on the interactive split of the WOMD.
\paragraph{\textbf{Ours} and \textbf{Ours guided}} These are our proposed consistency model trained and sampled with 4 steps, with and without the proposed guided sampling.

\subsubsection{\textbf{Training}}
We conduct distributed training with a total batch size of 100 across 4 NVIDIA L40 GPUs.
For consistency and diffusion models, we use the Adam optimizer~\cite{kingma2014adam} with a learning rate of $8\times10^{-5}$ and train for 50 epochs.
For the transformer, we follow the MTR settings~\cite{shi2022motion} and use the AdamW optimizer~\cite{kingma2014adam} with a learning rate of $1\times10^{-4}$ and weight decay of 0.01, and train for 30 epochs.

\subsection{Experiment results and analysis}

\subsubsection{\textbf{Interactive Behavior}}

We demonstrate that our predictive planner can generate safe and effective interactions with other road users with a single inference of the consistency model, without alternation between prediction and planning.

In Fig. \ref{fig: interactive behavior}, we show a proactive nudging and a safe yielding behavior of our planner, with the ego agent (``E") trajectory in green, the surrounding agents (``S") in orange, and other agents in blue.
In Fig.~\ref{fig:nudging}, our planner directs the ego vehicle to accelerate and nudge into the left lane, predicting that the vehicle in the left lane will decelerate to accommodate the lane change.
While in the ground truth, the rightmost vehicle eventually turns right instead of going straight, this discrepancy is expected for an 8-second future prediction.
In Fig.~\ref{fig:yielding}, the planner instructs the ego vehicle to wait until the pedestrian has fully crossed before making the right turn.

\subsubsection{\textbf{Planning Performance under Limited Computational Budget}}

\begin{figure*}[t]
  \centering
  \vspace{0.5em}
  \includegraphics[width=0.9\textwidth]{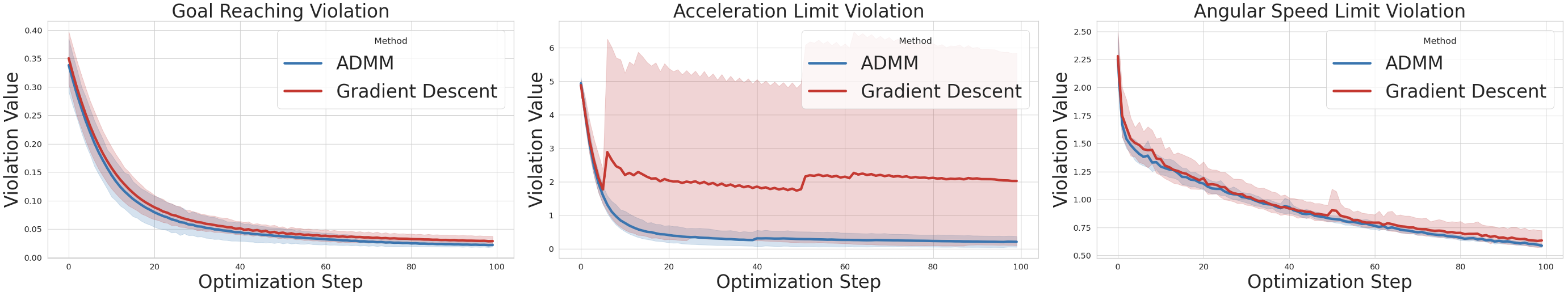}
  \caption{Constraint violation losses with respect to gradient steps during the final sampling step of the consistency model. 
  Vanilla gradient descent struggles with three conflicting constraints, while the ADMM-inspired method optimizes smoothly.}
  \label{fig:violation_loss_curve}
\end{figure*}


In Table \ref{tab:prediction metrics}, we quantitatively evaluate the realism of the planned trajectory for the ego vehicle through metrics with human groundtruth: Minimum Average Displacement Error (minADE) and Minimum Final Displacement Error (minFDE)~\cite{waymo2024motion} over 6 sampled trajectories.
The results show that our consistency model, with only 4 sampling steps, captures realistic human driving patterns comparable to Diffusion-10.
Moreover, the addition of guided sampling significantly improves the minFDE for the consistency models since it explicitly incorporates goal-reaching constraints.
We also report MotionDiffuser~\cite{shi2022motion} results on joint minADE and minFDE over the pair of interactive agents, which applies an attractor to the GT final point.
Our method achieves comparable performance to MotionDiffuser while requiring significantly fewer sampling steps (4 vs. 32).

In Table \ref{tab:trajectory quality and constraint violation}, we evaluate the trajectory quality using four metrics
1) Angle change, measuring the smoothness of directional transitions, 2) Path length, measuring the total distance traveled along the planning trajectory to reflect route efficiency, 3) Curvature, quantifying the sharpness of turns, and 4) Collision rate (with a $2m$ threshold), indicating safety.
Lower values in these metrics indicate smoother, safer, and more efficient trajectories that align with natural driving patterns.
Our consistency model achieves superior trajectory quality compared to all Diffusion-based methods, including Diffusion-10, despite its additional sampling steps.
While the transformer method has slightly lower angle changes, this is expected given its substantially larger model size (512 vs. our 128 dimensions).


Furthermore, we demonstrate the flexibility of our approach in choosing goal location in Fig. \ref{fig: different goal}.
Our consistency model maintains high-quality trajectory planning when targeting novel goal locations not present in the dataset.
The current transformer method doesn't support goal selections.

\subsubsection{\textbf{Ablation Study on ADMM-inspired Guided Sampling}}

We compare our ADMM-inspired method (ADMM for short) in Eq.~\eqref{eq: admm gradient step} with vanilla gradient descent (GD for short) in Eq.~\eqref{eq: original gradient descent}, which simultaneously optimizes all planning constraints.
To show the advantage of our method in hyperparameter tuning, we sweep through all combinations of step sizes ($\alpha = 1e^{-4}, 1e^{-5}, 1e^{-6}$) across three constraints (27 combinations total), using 50, 100, and 150 gradient steps over 200 scenarios. 
TABLE \ref{tab:admm_advantage} shows that GD results in significantly higher violations than ADMM on average; e.g., at 100 steps, GD has 9\% higher goal-reaching, 35\% higher acceleration, and 8\% higher angular speed violations.
With increasing gradient steps of 50, 100, and 150 steps, we obtain lower sums of violation as 1.45, 1.16, and 0.71, but with more computational tradeoffs.

In Fig. \ref{fig:violation_loss_curve}, we visualize how the violation loss evolves with ADMM (blue) and GD (red) methods in the last guided sampling step, measured across 200 scenarios and with the same step size finally chosen.
As discussed in Sec. \ref{sec: guided sampling}, these three constraints are in conflict with each other.
In Fig. \ref{fig:violation_loss_curve}, GD experiences large oscillations in acceleration violations (middle plot), highlighting its struggles with optimizing conflicting objectives.
In contrast, our ADMM method smoothly and effectively reduces all three violations.

\subsubsection{\textbf{Computational Time}}With the same encoder, the computational time for diffusion-based methods directly depends on the number of sampling steps, shown in Table \ref{tab:prediction metrics}. 
Diffusion-4 and DDIM have the same sampling step as our model, but fail to match our performance in trajectory generation across all metrics
in Table \ref{tab:prediction metrics} and Table \ref{tab:trajectory quality and constraint violation}.
The guided sampling, introducing a little computational overhead ($\approx 15\%$), provides higher quality trajectory generation and enables
controllable planning.
Even Diffusion-10, despite its more sampling steps, cannot achieve a comparable performance of trajectory generation to our approach in Table \ref{tab:trajectory quality and constraint violation}.



\section{Conclusion}

We propose a consistency model-based predictive planner that generates safe and highly interactive behavior, such as nudging and yielding, for autonomous vehicles.
Compared to diffusion models and their accelerated variants like DDIM, our method generates smoother, safer, and more efficient paths under a limited computational budget.
To satisfy multiple planning constraints online, we introduce a novel ADMM-inspired approach used in guided sampling for consistency models.
Future work will explore the integration of LLMs as high-level planners for closed-loop decision-making in long-tailed challenging scenarios.

\bibliographystyle{plain}
\bibliography{root.bbl}

\end{document}